\documentclass[letterpaper, 10 pt, conference]{ieeeconf}  

\usepackage{epsfig}
\usepackage{graphicx}
\usepackage{amsmath}
\usepackage{amssymb}
\usepackage{algorithm}
\usepackage{algpseudocode}
\usepackage{float}
\usepackage{color}
\usepackage{colortbl}
\usepackage{rotating}
\usepackage{epsfig}
\let\labelindent\relax
\usepackage{enumitem}
\usepackage{makecell}
\usepackage{tabularx}
\usepackage{xcolor}
\usepackage{multirow}
\usepackage{graphicx}
\usepackage{hhline}
\usepackage{textcomp,booktabs}
\usepackage{caption}
\usepackage{stmaryrd}
\usepackage[mathscr]{eucal}
\usepackage[export]{adjustbox}
\usepackage{comment}
\usepackage{multicol,lipsum}
\usepackage{amsfonts}
\usepackage{arydshln}
\usepackage{amsmath}

\usepackage{hyperref}
\usepackage{url}
\usepackage{tabularx}

\colorlet{custom_yellow}{green!10!orange}
\def\1{mathbb{1}}

\def\0{{\bf 0}}
\def\1{{\bf 1}}

\definecolor{purple}{rgb}{0.56,0.27,0.68}
\definecolor{red}{rgb}{0.95,0.4,0.4}
\definecolor{purered}{rgb}{1,0,0}
\definecolor{blue}{rgb}{0.4,0.4,0.95}
\definecolor{darkblue}{rgb}{0,0,0.8}
\definecolor{grey}{rgb}{0.6,0.6,0.6}
\definecolor{col1}{RGB}{232, 161, 148}
\definecolor{col11}{RGB}{255, 228, 228}
\definecolor{col2}{RGB}{148, 187, 232}
\definecolor{col33}{RGB}{206, 239, 255}
\definecolor{col3}{RGB}{233, 255, 245}
\definecolor{lightgrey}{rgb}{0.85,0.85,0.85}
\definecolor{lightlightgrey}{rgb}{0.9,0.9,0.9}
\definecolor{verylightBG}{rgb}{0.9,0.99,0.99}
\definecolor{darkgreen}{rgb}{0., 0.85, 0.5}

\graphicspath{{./figures/}}   

\DeclareMathOperator*{\argmin}{arg\!\min}

\usepackage[capitalize]{cleveref}
\crefname{section}{Sec.}{Secs.}
\Crefname{section}{Section}{Sections}
\Crefname{table}{Table}{Tables}
\crefname{table}{Tab.}{Tabs.}

\IEEEoverridecommandlockouts                              

\title{\LARGE \bf
Planning with Adaptive World Models for Autonomous Driving
}

\author{Arun Balajee Vasudevan, Neehar Peri, Jeff Schneider, Deva Ramanan \\ Carnegie Mellon University
}

\begin{document}

\maketitle
\thispagestyle{empty}
\pagestyle{empty}

\begin{abstract}
Motion planning is crucial for safe navigation in complex urban environments. Historically, motion planners (MPs) have been evaluated with procedurally-generated simulators like CARLA. However, such synthetic benchmarks do not capture real-world multi-agent interactions. nuPlan, a recently released MP benchmark, addresses this limitation by augmenting real-world driving logs with closed-loop simulation logic, effectively turning the fixed dataset into a reactive simulator. We analyze the characteristics of nuPlan's recorded logs and find that each city has its own unique driving behaviors, suggesting that robust planners must adapt to different environments. We learn to model such unique behaviors with BehaviorNet, a graph convolutional neural network (GCNN) that predicts {\em reactive} agent behaviors using features derived from recently-observed agent histories; intuitively, some aggressive agents may tailgate lead vehicles, while others may not. To model such phenomena, BehaviorNet predicts the parameters of an agent's motion controller rather than directly predicting its spacetime trajectory (as most forecasters do). Finally, we present AdaptiveDriver, a model-predictive control (MPC) based planner that unrolls different world models conditioned on BehaviorNet's predictions.  Our extensive experiments demonstrate that AdaptiveDriver achieves state-of-the-art results on the nuPlan closed-loop planning benchmark, improving over prior work by 2\% on Test-14 Hard R-CLS, and generalizes even when evaluated on never-before-seen cities. 
\end{abstract}

\begin{figure}[t]
\centering
\includegraphics[width=\linewidth]{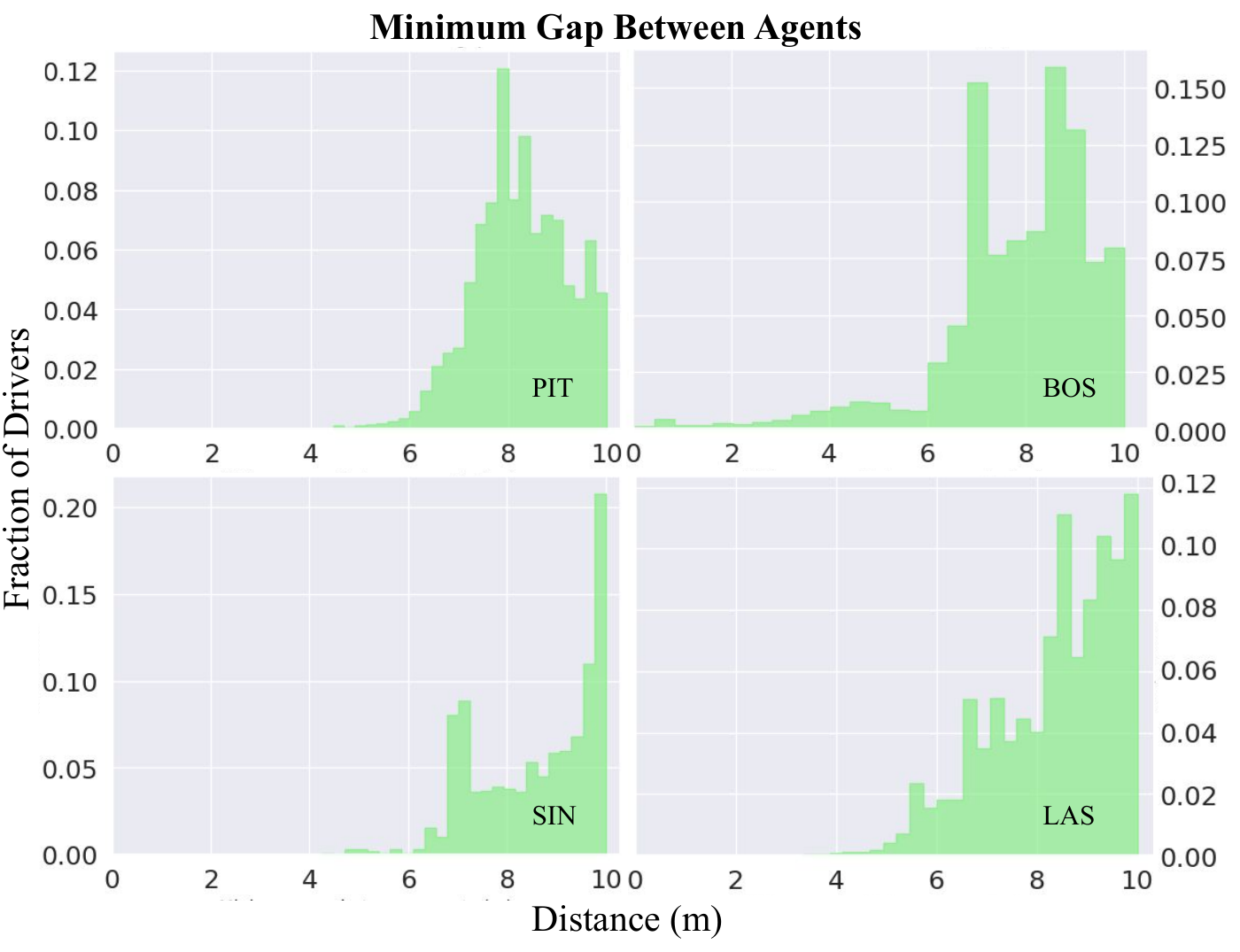}
\caption{\small {\bf Drivers Behave Differently in Different Cities.}
Using the nuPlan benchmark, we compare the distance between the ego-vehicle and lead agent (min-gap) in Pittsburgh (PIT), Boston (BOS), Las Vegas (LAS) and Singapore (SIN). Interestingly, we find that the min-gap distribution between cities differs dramatically. 
For example, Boston drivers are more aggressive (e.g. drive with lower average min-gap) than Pittsburgh drivers. This suggests that (a) robust planners must adapt to diverse driving conditions and (b) model-predictive control (MPC) planners may benefit from adaptive world models that capture such behaviors.}
\label{fig:teaser}
\end{figure}

\section{Introduction}
Motion planning (MP) is a critical component of the autonomy stack. Autonomous Vehicles (AVs) must carefully plan their motion to navigate in complex urban environments to safely reach their goal, avoid collisions, and abide by the rules of the road. Motion planners are typically trained and evaluated in synthetic environments like CARLA \cite{dosovitskiy2017carla} and AirSim \cite{airsim2017fsr}. However, such simulated environments notoriously suffer from a sim-to-real gap due to systematic biases and a lack of real-world diversity. Instead, we focus on the recent nuPlan planning benchmark~\cite{caesar2021nuplan}, which makes use of {\em real} driving logs to create environments for training and evaluation. Data-driven simulators like nuPlan are a core enabling technology in current autonomy stacks through the use of {\em re}simulation and log playback. 

{\bf Planning Evaluation.} Although nuPlan evaluates algorithms on ego-centric forecasting accuracy (defined as the OLS metric ~\cite{caesar2021nuplan}) and motion planning in a non-reactive world (NR-CLS), our work primarily focuses on motion planning in a reactive world (R-CLS) as this most closely resembles real-world deployment. Recent methods like PDM-C \cite{dauner2023parting} have dramatically improved planning performance in reactive environments by combining model-predictive control (MPC) with classical motion controllers like the Intelligent Driver Model (IDM) \cite{treiber2000congested}. 

{\bf Model Predictive Control.} Surprisingly, rule-based methods like PDM-C still outperform learning-based planners on real data. PDM-C builds upon IDM by incorporating several key ideas from the MPC literature like proposal generation, world model rollouts, and proposal scoring. First, PDM-C identifies the shortest spatial trajectory to the goal using a lane graph-based search. Next, it generates spatio-temporal proposals by applying IDM policies at different target speeds and lateral center-line offsets. Each proposal is evaluated in an internal world model and is scored based on traffic-rule compliance, progress towards the goal, and ego-vehicle comfort. The proposal with the highest score is selected and executed as the final motion plan. 

{\bf Different World Models for Each City.}
Intuitively, PDM-C could achieve perfect planning performance given an accurate world model of future agent actions and an infinite number of trajectory proposals. However, evaluating complex world models can be computationally infeasible. Instead, PDM-C makes use of a simpler “world-on-rails” model where agents move with constant velocity during rollouts. Notably, this non-reactive model does not accurately simulate agent behaviors, leading to poorly scored proposals. 
To improve this ``world-on-rails'' model, we analyze the characteristics of nuPlan's recorded logs and find that each city has its own unique driving behaviors. For example, Boston drivers are more aggressive (e.g. drive with lower average min-gap) than Pittsburgh drivers (cf. Fig. \ref{fig:teaser}).  We model such unique behaviors with BehaviorNet, a graph convolution-based network that predicts future agent actions based on features derived from recently observed agent behaviors. Unlike traditional forecasters, BehaviorNet parameterizes future agent behaviors with IDM controls to create a {\em reactive} world model. Conceptually, each set of predicted control parameters unrolls a different world model. 


{\bf Building an Adaptive Planner.} 
Finally, we introduce AdaptiveDriver, an MPC-based planner that scores spatio-temporal proposals conditioned on BehaviorNet's reactive world model. Extensive experiments demonstrate that our AdaptiveDriver planner achieves state-of-the-art results on the nuPlan closed-loop planning benchmark, improving by 2\% on Test-14 Hard. Importantly, the novelty of our work is in our parameterization of future agent behaviors as control parameters (rather than non-reactive spacetime forecasts), which implicitly encodes an ensemble of reactive world models that generalize to never-before-seen environments.

\textbf{Contributions.} We present three major contributions.

\begin{enumerate}[noitemsep,  topsep=-1pt]
    \item We demonstrate that each city has its own unique driving behaviors, and adapting to these different environments leads to notable improvements in planning performance. Moreover, we find that behaviors can even vary within a city, motivating our next contribution.
    \item We propose BehaviorNet, a graph convolutional neural network (GCNN) that predicts driving behaviors parameterized as IDM controls, using features derived from recently observed agents in the surrounding scene. 
    \item We present AdaptiveDriver, a planner based on model-predictive control (MPC) that unrolls and executes adaptive world models to safely navigate in diverse environments, achieving state-of-the-art closed-loop planning performance on nuPlan.

\end{enumerate}

\begin{figure}[t]
    \centering
    \includegraphics[width=\linewidth]{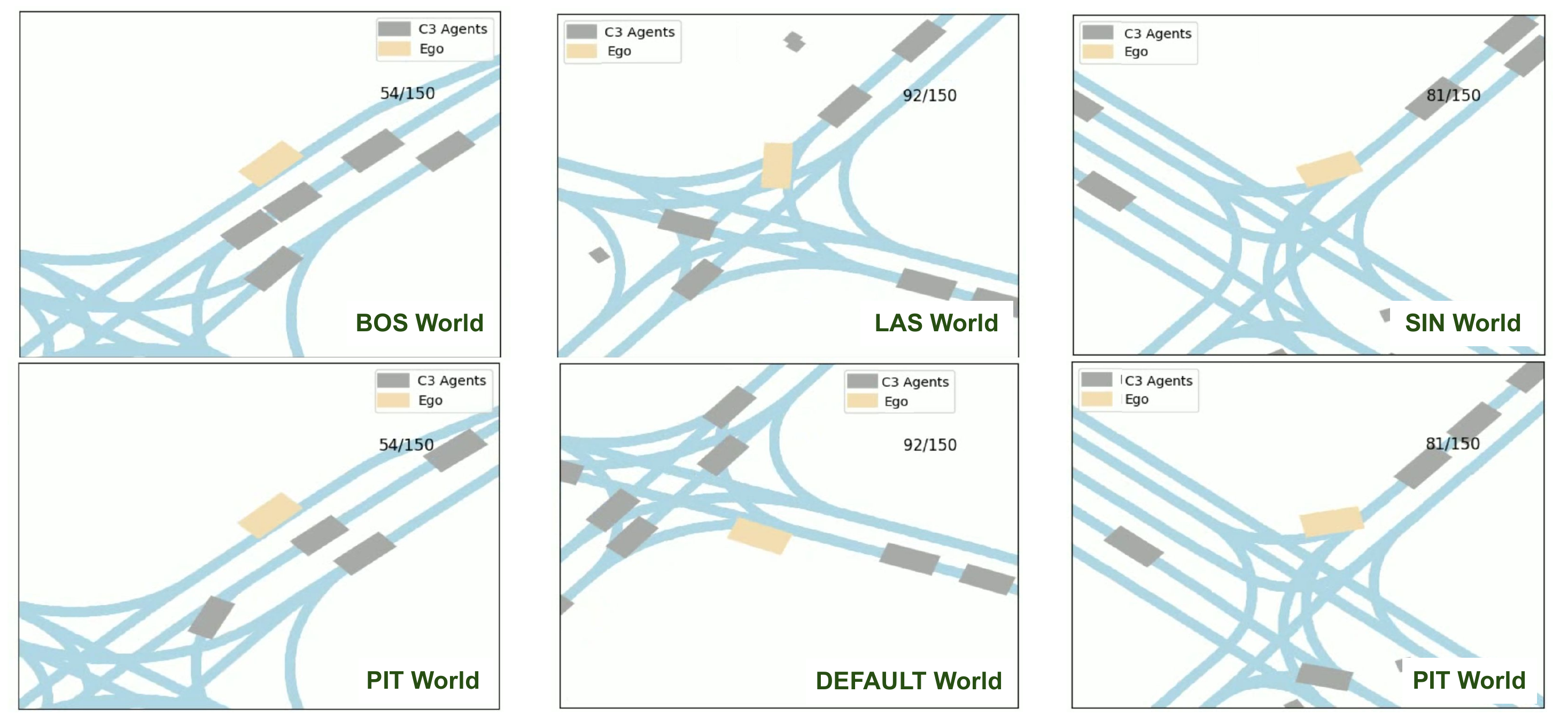}
    \caption{\small \textbf{Visualizing Adaptive World Models.} 
    Each column visualizes the same initial traffic scenario unrolled with different world models (trained on agents from different cities). We visualize the ego-vehicle in yellow and other agents in gray. Blue lines represent the lane-graph.
    On the {\bf left}, the BOS world model (top) produces agents that tend to tailgate more than the PIT world model, consistent with the min-gap statistics across those two datasets in Figure~\ref{fig:teaser}. In the {\bf center}, agents in the LAS world model have a higher max acceleration compared to those of the default IDM world model. On the {\bf right}, agents in the SIN world model have a higher max acceleration and min-gap compared to the PIT world model. We demonstrate that such adaptive world models can be used to significantly improve the accuracy of model-predictive control (MPC) planners. To do so, we train BehaviorNet, a network that predicts parameters of an adaptive world model based on the recent observed history of agents in the scene.} 
    \label{fig:city-behaviors}
\end{figure}

\section{Related Works}

\textbf{Rule-Based Planning.} Although recent works focus on learning robust policies by predicting goal-conditioned way-points \cite{rhinehart2019precog}, cost-volumes, and reward functions \cite{scheel2022urban}, rule-based planners still outperform learning-based approaches on real data \cite{dauner2023parting}. Rule-based planners are well studied ~\cite{stentz1994optimal,lavalle2001randomized,reeds1990optimal,gonzalez2015review,zhou2022review}, and have been widely adopted due to their safety guarantees and interpretability  \cite{thrun2006stanley, bacha2008odin, leonard2008perception, urmson2008autonomous, chen2015deepdriving}. Given the current position, velocity, and distance to the lead vehicle, rule-based planners estimate longitudinal acceleration to safely progress towards the target. The Intelligent Driver Model (IDM) \cite{treiber2000congested} is a classic non-learned algorithm for vehicle motion planning that relies on graph-based search to reach the target while employing a PID velocity controller to avoid collisions with other vehicles. Dauner et. al. \cite{dauner2023parting} upgrades IDM by sampling multiple trajectories and unrolling a constant velocity world model to select an optimal trajectory with minimum cost.

\textbf{Trajectory Optimization}. Motion planning is often framed as an optimization problem of hand-designed cost functions, which are then minimized to generate an optimal trajectory \cite{martin2009grandchallenge, montemerlo2008junior, fan2018baidu, ziegler2014julius}. To simplify this process, cost functions assume a quadratic objective function or divide the planning task into its lateral and longitudinal components. Approaches such as A* \cite{ajanovic2018search}, RRT \cite{karaman2011sampling}, and dynamic programming \cite{fan2018baidu} are commonly used to search for optimal solutions. CoverNet \cite{phan2020covernet} generates a set of trajectories and evaluates them based on cost functions, selecting the trajectory with lowest cost. While these methods are attractive due to their parallelizability, interpretability and functional guarantees, they are not robust when applied to real-world scenarios and require significant hyperparameter tuning. 
Conventional trajectory optimization approaches typically aim to compute a complete trajectory that spans from the initial configuration to the desired goal configuration. However, given the inherently dynamic and uncertain nature of the driving environment, precise long-horizon motion plans cannot be predicted in advance. As a result, model-predictive control (MPC) has gained prominence in recent years for real-time path planning \cite{rastelli2014dynamic, lavalle2006planning, karaman2010incremental, pongpunwattana2004real} because it adopts an iterative cost minimization strategy to select a locally optimal trajectory for each timestep. 

 \textbf{Data-Driven Simulation}. In recent years, many learning-based planners have emerged, leveraging the availability of simulator environments like CARLA \cite{dosovitskiy2017carla}, AirSim \cite{airsim2017fsr}, and others. However, current simulators are limited because they rely on synthetic data generated from game engines and have insufficient visual fidelity. Importantly, they lack the necessary diversity of driving scenarios required for comprehensive training and evaluation. To address these limitations, \cite{trafficsim} proposes multi-agent behavior models for generating diverse and realistic traffic simulations. More recently, \cite{montali2023waymo} introduced the Waymo sim agents challenge which evaluates simulators by comparing the trajectories of all simulated agents against their ground-truth trajectories.  Furthermore, CommonRoad \cite{commonroad} offers a driving dataset and planning benchmark which combines real-world data and rule-based heuristics.
 In contrast, nuPlan \cite{caesar2021nuplan} augments real-world driving logs with closed-loop simulation logic, effectively turning the fixed dataset into a reactive simulator. nuPlan has released $1300$ hours of real world driving logs from various cities including Las Vegas, Boston, Pittsburgh, and Singapore. Driving in each city presents a unique set of challenges. For example, Las Vegas has many high density pick-up and drop-off locations, and intersections with 8 parallel driving lanes per direction. In contrast, Boston drivers tend to double park, creating distinct planning challenges. 

\section{Planning With An Ensemble of Models}
\label{sec:city-specific-world-model}

In this section, we analyze the limitations of PDM-C~\cite{dauner2023parting} and propose AdaptiveDriver, an alternative instantiation of model-predictive control that achieves state-of-the-art closed-loop planning performance on the nuPlan benchmark~\cite{caesar2021nuplan}.

\begin{table}[t]
  \centering 
\begin{adjustbox}{max width=\linewidth,max totalheight=\textheight}
  \begin{tabular}{lcccccccc}
\toprule
   City  & Target Vel. ($\theta_{0}$) & Min Gap ($\theta_{1}$)  & Headway Time ($\theta_{2}$) & Max Acceleration ($\theta_{3}$) & Max Deceleration ($\theta_{4}$)\\ \midrule
 Default IDM World ($\theta^{IDM}$) & 10 & 1.0 & 1.5 & 1.0& 2.0 \\
 PIT World ($\theta^{PIT}$) & 10 & 2.0 & 0.5  & 1.5 & 3.0 \\ 
 BOS World ($\theta^{BOS}$)& 10 & 1.0 & 1.5 & 2.0 &  3.5 \\ 
 SIN World ($\theta^{SIN}$)& 10 & 2.5 & 1.5  & 2.0 & 1.0\\ 
 LAS World ($\theta^{LAS}$) & 10 & 1.0 & 0.5  & 1.5 & 0.5 \\ 
\bottomrule
\end{tabular} 
\end{adjustbox} 
\caption{\small \textbf{Optimized City-Specific Reactive World Models.} We generate city-specific IDM parameters ($\theta_0, \dots, \theta_4$) by optimizing the objective function described in Equation~\ref{eq:find_theta} over all training logs from a given city. The optimized world model parameters attempt to minimize the distribution shift between simulated trajectory rollouts and recorded logs on the training set.
}\label{tab:idm_param}%
\vspace{-1mm}
\end{table}

\textbf{nuPlan Evaluates Planners in a Reactive Simulator.} 
nuPlan augments real-world driving logs with closed-loop simulation logic, allowing other agents to react to the ego-vehicle. Agents are instantiated with an initial velocity based on their trajectory history and will re-simulate their spatial trajectory from the recorded driving log. The closed-world simulation logic for all agents is initialized with a fixed target velocity ($\theta_0$), minimum gap ($\theta_1$), headway time ($\theta_2$), maximum acceleration ($\theta_3$), and maximum deceleration ($\theta_4$). 

\textbf{Understanding PDM-C's Limitations}. PDM-C is a state-of-the-art rule-based planner that improves upon the Intelligent Driver Model (IDM), a car following algorithm that employs a simple longitudinal PID velocity controller along a reference path. PDM-C upgrades IDM to an MPC-based planner by modulating IDM's reference path with different longitudinal velocities and lateral offsets to generate candidate trajectories, internally unrolling a world model for other agents, and selecting the trajectory that minimizes a cost function over that world model. Notably, PDM-C makes use of a simpler ``world-on-rails" internal world model where other agents are non-reactive and move with constant velocity during rollouts. 
Although a `world-on-rails" model with constant velocity forecasts may work well for short-horizon forecasting, it fails to correctly simulate multi-agent interactions like lane changes and lane merges.

\begin{figure}[t]
    \centering
    \includegraphics[width=\linewidth]{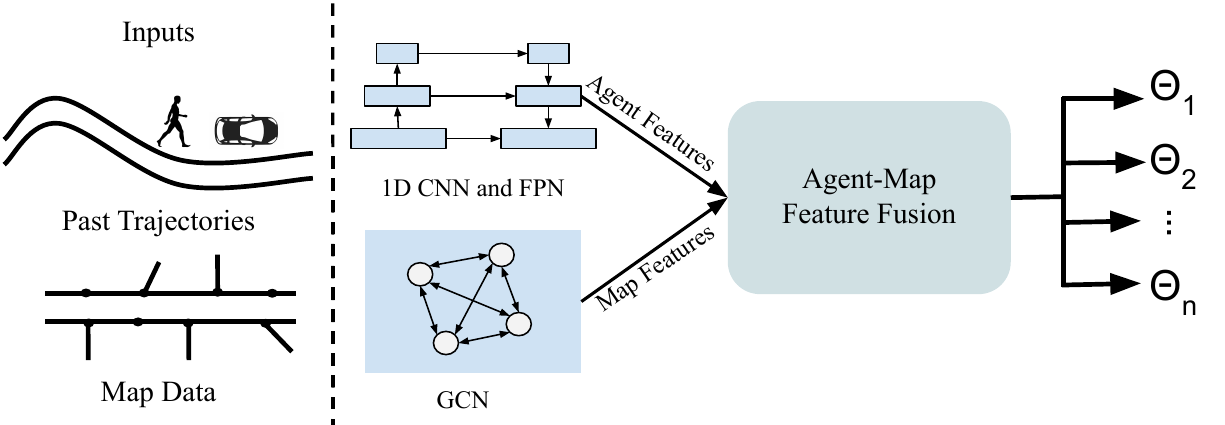}
    \caption{\textbf{BehaviorNet Architecture.} BehaviorNet uses past trajectories and map context to predict future agent behaviors parameterized as IDM contorls. Following LaneGCN \cite{liang2020learning}, we use a graph convolutional network (GCN) to extract map features from the lane graph. Next, we extract agent features from past trajectories. We then use LaneGCN's Agent-Map Feature Fusion to model interactions between agents and the map. Lastly, we pass these agent-map features through an MLP to predict IDM controls.}
    \label{fig:behaviornet_arch}
\end{figure}

\textbf{Predicting Future Agent Actions with BehaviorNet}.
We improve upon PDM-C's ``world-on-rails'' model by learning to predict future agent behaviors with BehaviorNet. We model the unique driving characteristics of each scenario by encoding a vectorized road graph of radius $R$ around the ego-vehicle and two seconds of trajectory history for all nearby agents. BehaviorNet consists of several multi-scale graph convolution and attention modules followed by a fully connected layer to predict IDM control parameters. We further describe BehaviorNet's architecture in Figure \ref{fig:behaviornet_arch}. Notably, unlike traditional forecasters, BehaviorNet directly predicts IDM control parameters ($\theta_0, \dots, \theta_4$), which can then used to unroll a reactive world model.

\textbf{Learning Adaptive Behavior Parameters}. We train BehaviorNet with paired examples of past agent trajectories and target IDM control parameters that best explain future agent actions. We optimize target IDM parameters by fitting to training logs using a grid search over $\boldsymbol\theta= \{\theta_0,\ldots,\theta_4\}$: 
\begin{equation}
    \boldsymbol\theta^{TARGET} = \argmin_{\boldsymbol\theta} ||X_{SIM}(\boldsymbol\theta) - X_{LOG}||_2^2
    \label{eq:find_theta}
\end{equation}
where $X_{SIM}$ is the position of all agents in the simulated IDM rollout and $X_{LOG}$ is the position of all agents in the recorded driving log. We assume that the ego-vehicle is non-reactive and simply replays its driving log. Intuitively, our optimization finds controls parameters that best tracks ground-truth agent positions. Importantly, Eq \ref{eq:find_theta} could be optimized over individual agents, over individual logs, or over all logs from a particular city. Motivated by our observation on city-specific behaviors, we first present results for optimizing target IDM parameters for all logs from a particular city (cf. Table~\ref{tab:idm_param}). We highlight salient observations below. 
First, $\boldsymbol\theta^{BOS}$ and $\theta^{LAS}$ tailgate vehicles more aggressively than $\boldsymbol\theta^{PIT}$ and $\boldsymbol\theta^{SIN}$, which better tracks the city-specific min-gap statistics from Figure~\ref{fig:teaser}.
Second, simulated agents using the default IDM world model ($\boldsymbol\theta^{IDM}$, optimized to minimize Eq. \ref{eq:find_theta} over the entire dataset) drive less aggressively with lower \emph{maximum acceleration} and \emph{maximum deceleration} than city-specific models from PIT ($\boldsymbol\theta^{PIT}$) and BOS ($\boldsymbol\theta^{BOS}$). Third, $\boldsymbol\theta^{BOS}$ seems to have the most aggressive drivers and $\boldsymbol\theta^{LAS}$ seems to have least aggressive ones, as defined by \emph{maximum acceleration} and \emph{maximum deceleration}. 


\begin{figure}[t]
\centering
\includegraphics[width=\linewidth]{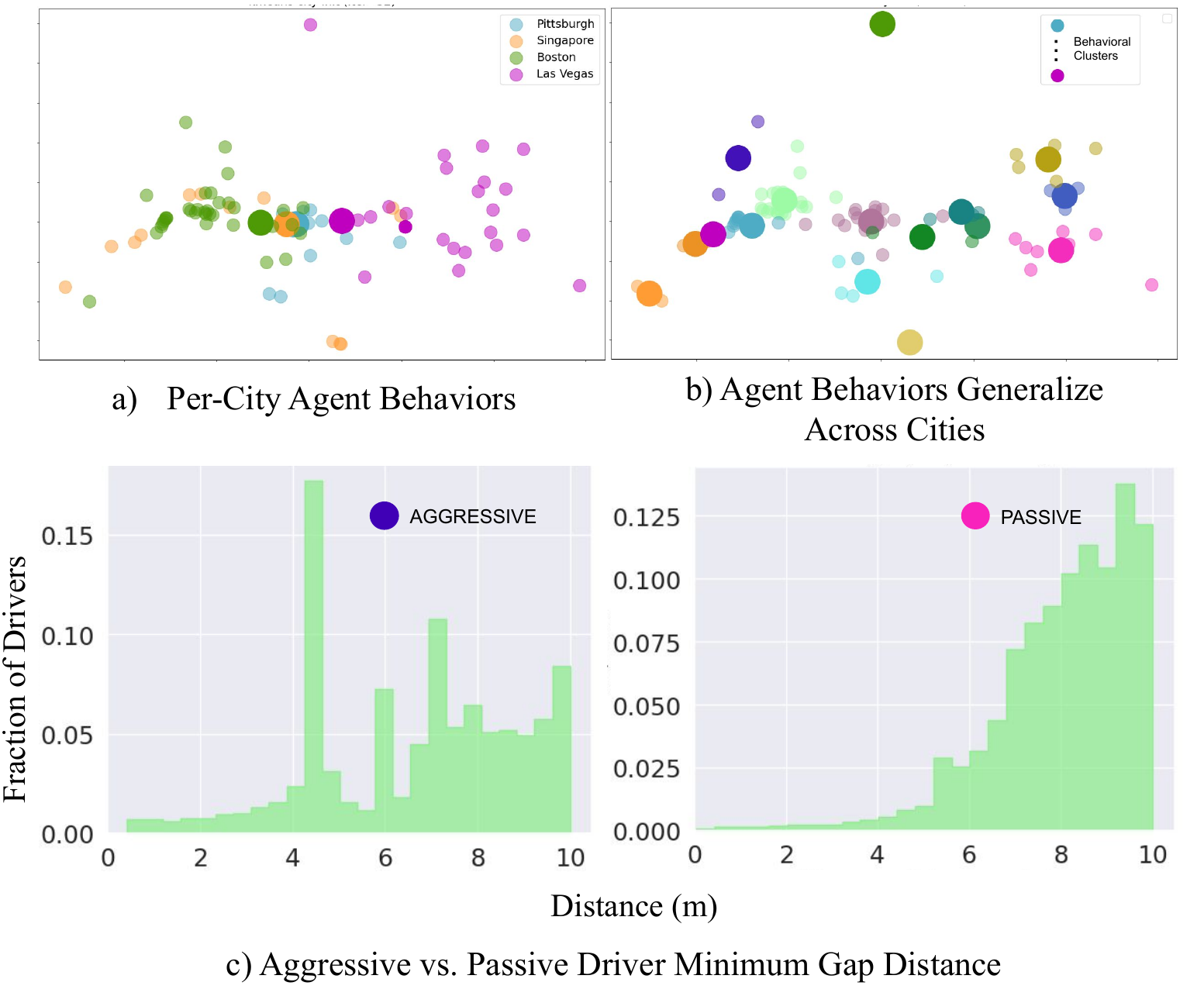}
\caption{\small {\bf Visualizing Clusters of Behaviors.}
In (a), we optimize scenario-specific world models using Equation~\ref{eq:find_theta} and visualize per-log IDM parameters using a tSNE plot, coloring cities differently. For comparison, we also plot per-city-optimized IDM parameters (Table~\ref{tab:idm_param}) as large colored dots. In (b), we cluster per-scenario IDM parameters using $K$-means and visualize different clusters. Each cluster represents a unique emergent driving behavior. In (c), we compare the distribution of min-gap between the ``aggressive driver'' cluster (left) and ``passive driver'' cluster (right). We note that the ``aggressive driver'' cluster has a lower average min-gap than the ``passive driver'' cluster, validating that our $K$-means clusters encode unique city-agnostic driving behaviors.}
\vspace{-3mm}
\label{fig:clustering}
\end{figure}

\textbf{Training Log-BehaviorNet.}
Although each city has distinct driving characteristics, 
agents can still behave differently within a city. For example, Boston drivers may tailgate within the city but drive more cautiously on highways. 
To model this, we simply optimize Eq.\ref{eq:find_theta} over each individual training log.
Figure~\ref{fig:clustering} (a) visualizes the set of log-specific IDM parameters $\{\boldsymbol\theta\}$ with tSNE, color-coded by city. Rather than training BehaviorNet to directly regress these parameters, we reframe the problem as a simpler discrete classification task. Specifically, we cluster the set of $\{ \boldsymbol\theta \}$ from the train-set into $K$ clusters and train BehaviorNet with a $K$-way softmax loss. We refer to this network as Log-BehaviorNet, in contrast to City-BehaviorNet.
Figure~\ref{fig:clustering}-(b) compares the learned behavior clusters with the original city ``clusters" in (a). Figure~\ref{fig:clustering}-(c) plots the min-gap distribution from two distinct clusters, suggesting that each cluster loosely corresponds to prototypical behaviors such as ``aggressive" or ``passive". 
We tune the number of behavior clusters ($K$) so as to maximize $R-CLS$ performance on the nuPlan train-set. Interestingly, the optimal number (16) is far more than the number of distinct cities (4). Importantly, we show that city-agnostic clusters generalize better than city-specific models, particularly when evaluated on never-before-seen cities (cf. Table \ref{tab:driver-generalization}). 

\begin{figure}[t]
    \centering
    \includegraphics[width=\linewidth]{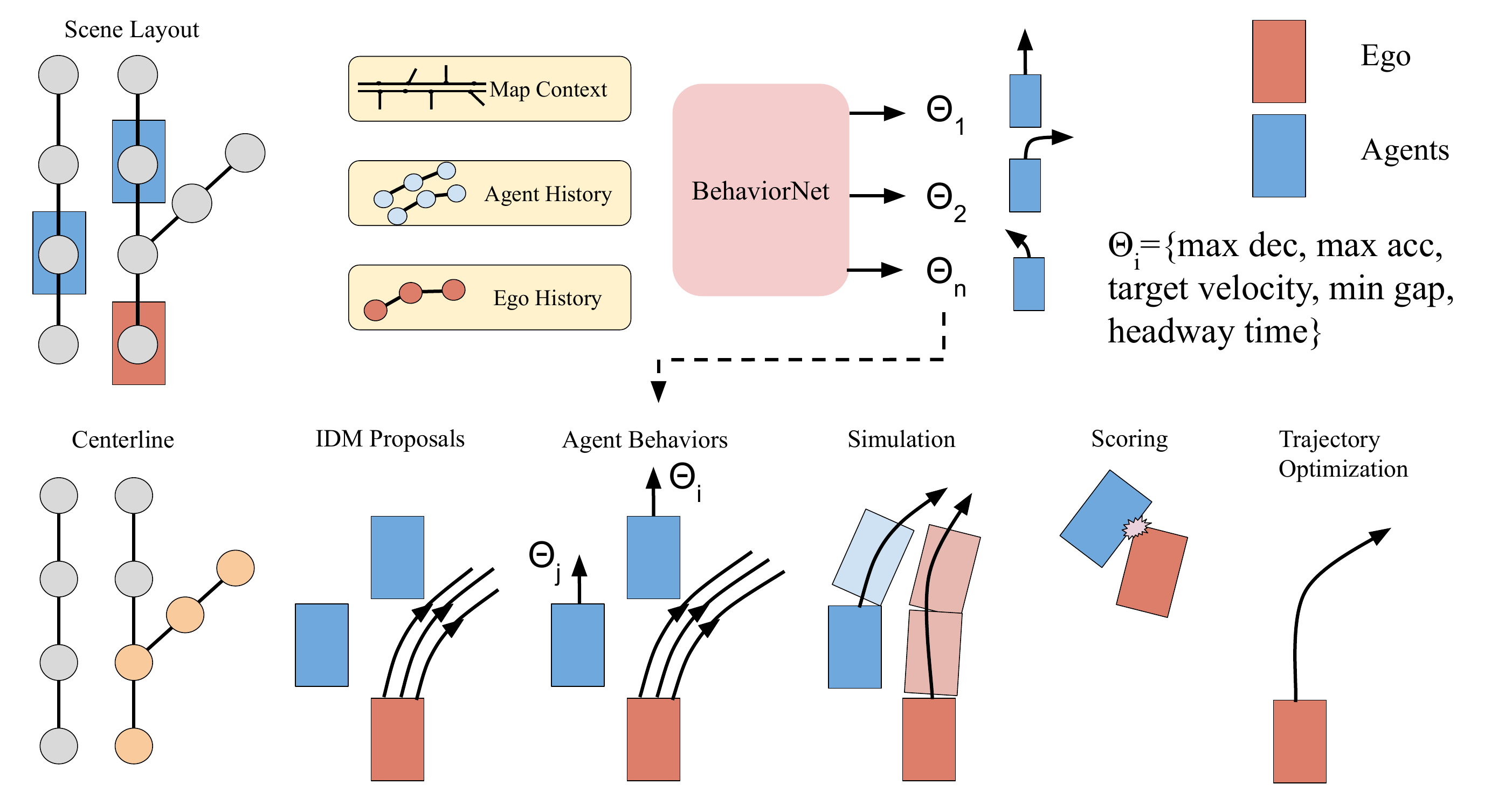}
    \caption{\small {\bf Overview of AdaptiveDriver's Architecture.} AdpativeDriver's architecture extends PDM-C by replacing ``world-on-rails' rollouts with adaptive reactive world models using BehaviorNet. We predict future agent behaviors parameterized as IDM controls using scene context including the ego-vehicle's history, past agent trajectories, and surrounding lane graph. Similar to PDM-C, our planner identifies the nearest center-line to the goal using graph-based search. We generate many trajectory proposals and score each according to BehaviorNet's reactive world model. The proposal with the highest score is selected and executed.
    }   
    \label{fig:architecture}
\end{figure}


\textbf{Adding Learned Priors into Rule-Based Planners}. Although rule-based planners like PDM-C still outperform learning-based methods on real data, they fail to accurately model future agent behaviors in world model rollouts. We aim to bridge the gap between rule-based and learning-based planners with AdaptiveDriver (cf. Fig. \ref{fig:architecture}), a model-predictive control (MPC) planner that uses behavior parameter predictions to improve the quality of world model rollouts.  Notably, although AdaptiveDriver and PDM-C are both instantations of MPC-based planners, our model improves upon PDM-C using (1) a reactive world-model that (2) adapts to each log using features derived from past agent behaviors. 

{
\setlength{\tabcolsep}{8mm}
\begin{table*}[t]
  \centering 
\begin{adjustbox}{max width=\linewidth,max totalheight=0.85\textheight}
  \begin{tabular}{lcc|cc}
\toprule
          &  \multicolumn{2}{c}{Val-14}& \multicolumn{2}{c}{Test-14 Hard}  \\ 
   Model  &  NR-CLS &  R-CLS & NR-CLS &  R-CLS  \\ \midrule
    Log Replay & 94.03    &  75.86   &  85.96   & 68.80    \\ 
    \midrule
    IDM ~\cite{treiber2000congested} & 70.39    & 72.42    & 56.16    &  62.26   \\ 
    PDM-C ~\cite{dauner2023parting} &  92.51   & 91.79    &  65.07   &  75.18   \\ 
    RasterModel ~\cite{djuric2020uncertainty} & 69.66    & 67.54    &  49.47   & 52.16    \\ 
    UrbanDriver ~\cite{scheel2022urban} &  63.27   & 61.02    &  51.54   & 49.07    \\ 
    GC-PGP ~\cite{hallgarten2023prediction} & 55.99    &  51.39   &  43.22   & 39.63    \\ 
    PDM-O ~\cite{dauner2023parting} & 52.80    & 57.23    &  33.51   &  35.83   \\ 
    GameFormer ~\cite{huang2023gameformer} & 80.80    & 79.31    &  66.59   &  68.83   \\ 
    PlanTF ~\cite{jcheng2023plantf} &  84.83   & 76.78    &  72.68   &  61.70   \\ 
    DTPP ~\cite{huang2023dtpp} & 89.64    &  89.78   &  59.44   &  62.94   \\ 
    LLM-Assist$_{UNC}$ ~\cite{sharan2023llmassist} & 90.11    &  90.32   &  -   &  -   \\ 
    LLM-Assist$_{PAR}$ ~\cite{sharan2023llmassist} &  93.05   & 92.20    &  -   &   -  \\ 
    PlanAgent ~\cite{zheng2024planagent} &  \textbf{93.26}   &  92.75   & 72.51    &  76.82   \\ 
  \midrule
  AdaptiveDriver w/ City-BehaviorNet (Ours) &  92.97 & 93.28 & 70.11 & 78.02 \\
  AdaptiveDriver w/ Log-BehaviorNet (Ours)&  93.15   & \textbf{93.49}    & \textbf{73.25}    &  \textbf{78.87}   \\ 
  
\bottomrule
\end{tabular} 
\end{adjustbox} 
\caption{\small \textbf{nuPlan Benchmark Results}. We evaluate several rule-based and learning-based planners on the nuPlan test set. First, we note that rule-based planners like IDM and PDM-C outperform prior learning-based methods like Raster Model, LaneGCN, and UrbanDriver on NR-CLS and R-CLS. In contrast, AdaptiveDriver demonstrates that carefully combining learning-based behavior priors with rule-based planners significantly improves over prior work by 2\% on Test14-Hard R-CLS.  Results of prior work are reproduced from \cite{jcheng2023plantf}. We provide additional visuals comparing our planner to our PDM-C baseline in Fig \ref{fig:qualitative}.
}\label{tab:benchmark}%
\end{table*}
}

\section{Experiments}
In this section, we compare AdaptiveDriver with several baselines on the nuPlan benchmark and show that our proposed approach achieves state-of-the-art closed-loop planning performance. Further, we demonstrate that AdaptiveDriver generalizes to never-before-seen cities. 

\textbf{nuPlan Dataset}. We evaluate all methods on the nuPlan benchmark~\cite{caesar2021nuplan}, which includes $\sim$10M driving logs,  collected in Boston, Pittsburgh, Las Vegas and Singapore. nuPlan identifies $73$ types of interesting scenarios e.g. \{{\tt changing lanes, starting left turn, unprotected right turn, etc.}\}, but only evaluates $14$ scenarios in the official benchmark. We evaluate on the official Val-14 and Test-14 Hard \cite{jcheng2023plantf} splits in Table \ref{tab:benchmark}, and evaluate all ablations on a representative mini val-set. We construct the mini val-set by sampling 10 scenarios from all 14 scenario types across all 4 cities (560 logs). 

\textbf{Evaluation Setup and Metrics}. The nuPlan benchmark evaluates planners under three evaluation settings: OLS, NR-CLS and R-CLS. Following prior work \cite{jcheng2023plantf, yang2024diffusion,zheng2024planagent}, we focus on NR-CLS and R-CLS as these metrics directly evaluate planning-centric performance. Specifically, \emph{NR-CLS} and \emph{R-CLS} measure \emph{ego progress along expert trajectory}, \emph{speed limit compliance}, \emph{driving direction compliance}, \emph{time to collision within bounds}, and \emph{ego is comfortable}. The overall score is computed using a weighted sum of the above metrics. We refer the reader to  ~\cite{caesar2021nuplan} for further details.

\textbf{Implementation Details.} We build upon the codebases of PDM-C~\cite{dauner2023parting} and nuPlan~\cite{caesar2021nuplan} for this work. AdaptiveDriver adds an IDM controller on top of PDM-C's~\cite{dauner2023parting} constant velocity world model rollouts. 
We train BehaviorNet for $10$ epochs using the Adam optimizer~\cite{kingma2014adam} with a learning rate of $5\mathrm{e}{-5}$. We select a map context of radius $R=100$m. Our code is available on \href{https://arunbalajeev.github.io/world_models_planning/world_model_paper.html}{GitHub}.


{
\setlength{\tabcolsep}{8mm}
\begin{table}[t]
  \centering 
\begin{adjustbox}{max width=\linewidth,max totalheight=0.85\textheight}
  \begin{tabular}{llccccccc}
\toprule
           Model  & City  & NR-CLS  & R-CLS\\ \midrule
        PDM-C~\cite{dauner2023parting}   & PIT  & 92.94 & 90.59 \\ 
            & BOS  & 95.56 & 94.45 \\ 
            & SIN  & 93.49 & 94.15 \\ 
            & LAS  & 95.95 & 95.21  \\  \midrule
         AdaptiveDriver w/ City-BehaviorNet (Ours) & PIT   & 91.48 & 91.94 \\ 
          & BOS    & 96.32 & 96.21 \\ 
           & SIN    & 95.31 & 95.29 \\ 
           & LAS   & 96.75 & 97.08 \\ \midrule
         AdaptiveDriver w/ Log-BehaviorNet (Ours)& PIT   & 92.39 & 92.42 \\ 
         & BOS    & 96.39 & 96.44 \\ 
          & SIN    & 95.22 & 95.30 \\ 
           & LAS   & 96.61 & 97.25 \\
        \bottomrule
\end{tabular} 
\end{adjustbox} 
\caption{\small \textbf{Per-City Breakdown Analysis}. We observe that all planners with reactive world models (e.g. AdaptiveDriver w/ City-BehaviorNet and AdaptiveDriver w/ Log-BehaviorNet) consistently outperform PDM-C in closed-loop (e.g. reactive) planning performance (R-CLS) on the mini val-set. However, reactive world models do not consistently improve open-loop (e.g. non-reactive) planning performance (NR-CLS).}
\label{tab:city_specific_driver}
\end{table}
}

\textbf{Comparison to State-of-the-Art}. We evaluate several rule-based and learning-based planners on the nuPlan test set. First, we find that AdaptiveDriver w/ Log-BehaviorNet showcases 1.7\% and 3.7\% improvement on R-CLS over the PDM-C baseline on Val-14 and Test-14 Hard respectively. Notably, our approach outperforms more complex LLM-based planers \cite{sharan2023llmassist}, and convincingly outperforms our PDM-C baseline on Test-14 Hard. We present a per-city breakdown analysis in Table \ref{tab:city_specific_driver}. Interestingly, we find that all planners with reactive world models consistently outperform the baseline in closed-loop planning performance (R-CLS), with improvements in performance of up to $1.74\%$. Further, we confirm the (obvious fact) that motion planning for AVs is more difficult in some cities than in others. For example, although we achieve more than 97\% closed-loop planning accuracy in LAS, we only achieve 92\% performance in PIT.

{
\setlength{\tabcolsep}{8mm}
\begin{table}[!tb]
  \centering 
\begin{adjustbox}{max width=\linewidth,max totalheight=\textheight}
  \begin{tabular}{llccccccc}
\toprule
   Model  & City & NR-CLS &  R-CLS  \\ \midrule
  AdaptiveDriver w/ City-Oracle & All & 95.10 & 95.20 \\ 
  AdaptiveDriver w/ City-BehaviorNet & All & 94.97 ({\small \textcolor{red}{-0.13}}) & 95.13 ({\small \textcolor{red}{-0.07}}) \\
    \midrule
   AdaptiveDriver w/ Log-Oracle & All & 95.29  & 95.47 \\
  AdaptiveDriver w/ Log-BehaviorNet & All & 95.15 ({\small \textcolor{red}{-0.14}})  & 95.35 ({\small \textcolor{red}{-0.12}}) \\

\bottomrule
\end{tabular} 
\end{adjustbox} 
\caption{\small \textbf{Oracle Target IDM Parameters}. We evaluate the quality of BehaviorNet's predictions compared to oracle target IDM parameters on the mini val-set. Specifically, City-Oracle and Log-Oracle always use the target IDM parameters (rather than predicted IDM parameters) that minimizes Eq. \ref{eq:find_theta} for each log in world model rollouts. Notably, BehaviorNet closely matches the performance of the oracle on the test set, suggesting that our model accurately estimates IDM control parameters from scene context. 
}\label{tab:oracle_ablation}%
\vspace{-2mm}
\end{table}
}

{
\setlength{\tabcolsep}{8mm}
\begin{table}[t]
  \centering 
\begin{adjustbox}{max width=\linewidth,max totalheight=0.85\textheight}
  \begin{tabular}{llccccccc}
        \toprule
           Train Cities & Test City & NR-CLS &  R-CLS \\ \midrule
         PIT, BOS, SIN & LAS  & 96.55 ({\small \textcolor{red}{-0.06}})  & 97.21 ({\small \textcolor{red}{-0.04}}) \\ 
         All           & LAS  &  96.61      &   97.25   \\
         \midrule
         PIT, BOS, LAS & SIN  &  95.05 ({\small \textcolor{red}{-0.17}}) & 95.12 ({\small \textcolor{red}{-0.08}})\\ 
         All           & SIN  &   95.22     &   95.20   \\
         \midrule
        
         BOS, SIN, LAS & PIT  & 91.60 ({\small \textcolor{red}{-0.79}})  & 92.02 ({\small \textcolor{red}{-0.40}})\\
         All           & PIT  & 92.39       &   92.42   \\
         \midrule
        
         PIT, SIN, LAS & BOS  & 96.22 ({\small \textcolor{red}{-0.17}})  & 96.23 ({\small \textcolor{red}{-0.21}}) \\ 
         All           & BOS  & 96.39       &   96.44   \\
        
        \bottomrule
\end{tabular} \vspace{-2mm}
\end{adjustbox} 
\caption{\small \textbf{AdaptiveDriver Generalizes to New Cities}. Although learning-based planners can leverage data-driven priors, such methods may fail when evaluated in out-of-distribution scenarios. We evaluate AdaptiveDriver's ability to generalize to previously unseen cities in the mini val-set by holding out city-specific sequences when training BehaviorNet. We find that the planning performances (NR-CLS and R-CLS) are competitive with the model trained on all cities, indicating AdaptiveDriver's generalizability.
}\label{tab:driver-generalization}%
\end{table}
}

\begin{figure}[b]
    \centering
    \includegraphics[width=0.49\linewidth]{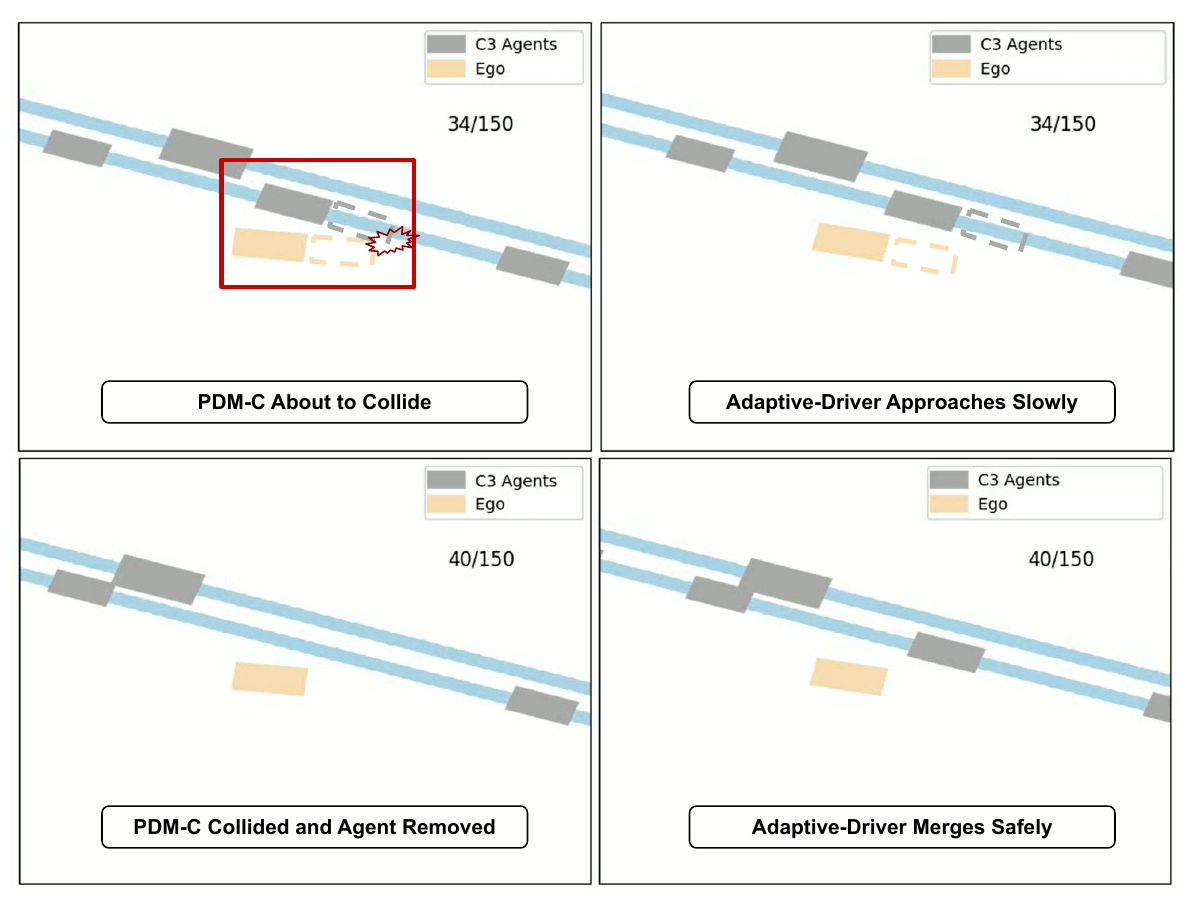} 
    \includegraphics[width=0.49\linewidth]{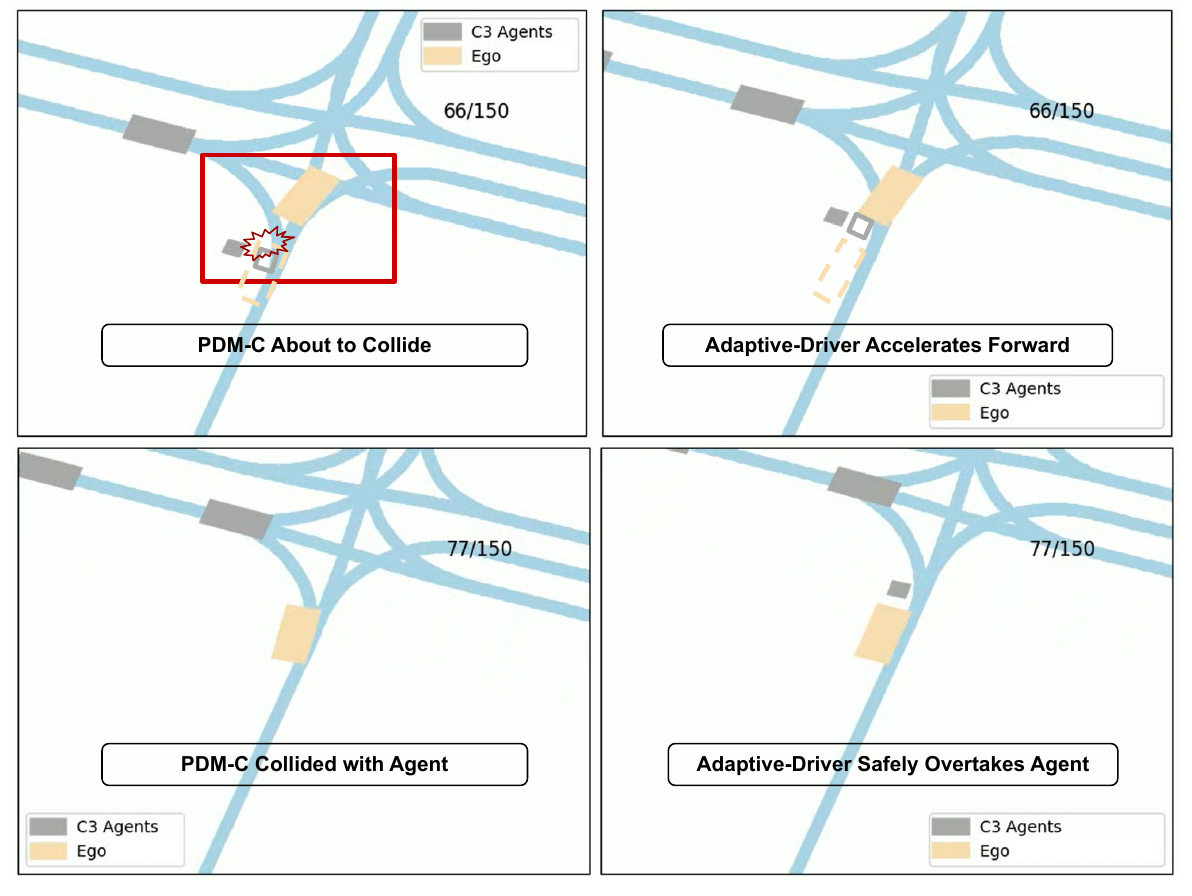}
    \caption{\small \textbf{Qualitative Comparison of AdaptiveDriver vs. PDM-C}. We evaluate the performance of our AdaptiveDriver against the current state-of-the-art in the simulation of two test scenarios (left and right) from the nuPlan dataset. These scenarios are 15 seconds long and sampled at 10 Hz. Timestamps are displayed on both the top and bottom rows to illustrate collision events. The yellow box represents the ego vehicle and the gray boxes represent other agents. According to nuPlan's simulation logic, vehicles that collide with each other are removed from the simulation. On the left, the ego vehicle moves from a parked position onto the road. AdaptiveDriver initially slows down to avoid a collision before smoothly entering the road, while PDM-C collides. On the right, we observe an agent about to cross the road. PDM-C drives slowly and collides, whereas our planner safely navigates forward. Please see our \href{https://arunbalajeev.github.io/world_models_planning/world_model_paper.html}{project page} for videos. 
    }
    \label{fig:qualitative}
\end{figure}

\textbf{Oracle Target Parameters vs. BehaviorNet Predicted Parameters}. 
 We assess the quality of BehaviorNet's predictions by comparing them to target IDM parameters from Eq. \ref{eq:find_theta}. The planning performance of AdaptiveDriver w/ Log-BehaviorNet closely approximates that of BehaviorNet w/ Log-Oracle, with differences of $0.14$ and $0.12$ in NR-CLS and R-CLS, respectively. The minimal difference between AdaptiveDriver using City-Oracle vs. City-BehaviorNet, and Log-Oracle vs. Log-BehaviorNet in Table \ref{tab:oracle_ablation}, suggests that our approach can reliably predict future agent actions.

\textbf{AdaptiveDriver Generalizes to New Cities}. As illustrated in Figure~\ref{fig:clustering} (b), we observe that many agent behaviors are shared across cities. To validate this observation, we train AdaptiveDriver w/ Log-BehaviorNet on three cities and test it on a held-out city (cf. Table~\ref{tab:driver-generalization}). We find that the planning scores on NR-CLS and R-CLS for seen-vs-unseen cities are comparable, suggesting that AdaptiveDriver can generalize beyond city boundaries.

\section{Conclusion}
In this paper, we demonstrate that each city has its own unique driving behaviors (e.g. Boston drivers tend to tailgate more than Pittsburgh drivers) and learn to model unique driving characteristics with BehaviorNet. We propose AdaptiveDriver, a model-predictive control (MPC) that unrolls and executes behavior-specific world models conditioned on BehaviorNet's predictions and achieves state-of-the-art performance on the nuPlan closed-loop reactive benchmark.  


\textbf{When does AdaptiveDriver Fail?} 
Since our world models build on IDM's PID controller, we inherit its flaws. We note that IDM can be too conservative (when it mistakes a parked vehicle for a lead vehicle and stops) or too aggressive (when traveling at high speeds along a curve. Secondly, IDM only interacts with the lead vehicle, limiting multi-agent reasoning.  Although we primarily focus on building city-specific and log-specific world models using the IDM, modeling agent-specific behavior models can potentially yield better performance. Future work should explore alternative optimization functions to maximize distribution similarity between real driving logs and simulated rollouts for learning control parameters. We include additional videos on our project page.

\textbf{Why Directly Predict Control Parameters?} Directly predicting control parameters allows our world model to more accurately simulate multi-agent interactions in rollouts, improving proposal scoring. Further, relying on a motion controller allows us to more easily capture nuanced behaviors that may be difficult to learn like speeding up when merging in rollouts. However, relying on a rule-based controller for rollouts also implies that we inherit its limitations. Moreover, unlike learning-based approaches that directly predict agent behaviors, our approach may not significantly improve with more data. Given enough data and the right learning objectives, we posit that methods that directly predict agent behaviors may be able to mimic the characteristics of MPC-based controllers.

\textbf{BehaviorNet Rollout Accuracy vs. Runtime} BehaviorNet is limited by IDM’s lateral capability, preventing it from simulating multi-agent interactions like lane changes and lane merges. In theory, one could use more sophisticated controllers that allow for lang changes and merges that differ from the reference path (like the controller in PDM-C).  Alternatively, one could replace a motion controller with an ego-forecaster such as MTR++ \cite{shi2024mtr++} or QCNet \cite{Zhou_2023_CVPR}. We believe both directions are fruitful paths forward for building better world models. In practice, we find that a simple IDM-based world model already achieves state-of-the-art performance. Moreover, similar to PDM-C’s constant velocity world model, our rule-based rollouts allow AdaptiveDriver to operate with lower latency. In contrast, running a SOTA trajectory prediction model at every timestep will significantly increase the latency of proposal scoring and is likely prohibitively slow for practical applications. 

\textbf{Learning-Based Rollouts Have Higher Latency}
We benchmark the latency of PDM-C and AdaptiveDriver on 1 RTX 3090 GPU with a batch size of 1 in Table \ref{tab:latency}. Notably, BehaviorNet only increases AdaptiveDriver’s runtime by 17ms compared to PDM-C. 

{
\setlength{\tabcolsep}{18mm}
\begin{table}[t]
  \centering 
\begin{adjustbox}{max width=\linewidth,max totalheight=0.95\textheight}
  \begin{tabular}{llcccccccc}
\toprule
   Model   &  Latency (ms)  \\ \midrule

  PDM-C &  232 \\
  AdaptiveDriver w/ Log-BehaviorNet  & 249 \\
  
\bottomrule
\end{tabular} 
\end{adjustbox} \vspace{1.5mm}
\caption{\small \textbf{Latency Analysis.} We compare the latency of PDM-C and AdaptiveDriver w/ Log-BehaviorNet. Unsurprisingly, AdaptiveDriver is slower than PDM-C due to our learning-based rollouts. 
}\label{tab:latency}%
\end{table}
}

\section{Acknowledgements}
This work was supported in part by funding from Bosch Research and the NSF GRFP (Grant No. DGE2140739).

\bibliographystyle{IEEEtran}
\bibliography{egbib.bib}

\end{document}